# Per-pixel Classification Rebar Exposures in Bridge Eye-inspection

Yasuno Takato[*1]  Nakajima Michihiro[*1]  Noda Kazuhiro[*1]

[*1]Yachiyo Engineering, Co., Ltd.

Efficient inspection and accurate diagnosis are required for civil infrastructures with 50 years since completion. Especially in municipalities, the shortage of technical staff and budget constraints on repair expenses have become a critical problem. If we can detect damaged photos automatically per-pixels from the record of the inspection record in addition to the 5-step judgment and countermeasure classification of eye-inspection vision, then it is possible that countermeasure information can be provided more flexibly, whether we need to repair and how large the expose of damage interest. A piece of damage photo is often sparse as long as it is not zoomed around damage, exactly the range where the detection target is photographed, is at most only 1%. Generally speaking, rebar exposure is frequently occurred, and there are many opportunities to judge repair measure. In this paper, we propose three damage detection methods of transfer learning which enables semantic segmentation in an image with low pixels using damaged photos of human eye-inspection. Also, we tried to create a deep convolutional network from scratch with the preprocessing that random crops with rotations are generated. In fact, we show the results applied this method using the 208 rebar exposed images on the 106 real-world bridges. Finally, future tasks of damage detection modeling are mentioned (211words).

## 1. Introduction

Deterioration of civil engineering structures is progressing in recent years, including a large number of concrete structures. Improving efficiency of scheduled inspections is a pressing issue, since the cost of inspections comprises a large proportion of maintenance costs for local governments, which are also experiencing manpower shortage for technical personnel. There are often opportunities to apply deep learning as a method for improving efficiency of inspections on social infrastructure and studies have been conducted on this issue. Close eye-base inspection is required for bridges once every five years and as a result, images of damage have been accumulating (Ministry of Land, Infrastructure, Transport and Tourism, 2014). If it were possible to utilize images of damage that are attached to inspection reports, data from scheduled inspections from past years can be input for the purpose of deterioration learning. If it could be possible to automatically calculate numerical scores for the extent of damage based on images of damage in addition to the conventional five-level qualitative evaluation, this would be useful in deciding whether any repairs work should be performed and for setting the order of priority among candidates for repairs. There are past studies on detecting cracks in concrete on bridges, structures, plants, etc. Cracks have a high rate of occurrence and it is relatively easy to produce annotation images. On the other hand, separation and rebar exposure progress to rebar corrosion, and are therefore considered to have greater impact on the health of structures. The detection model for separation and rebar exposure, however, is only at its incipient stages, and as such, it would be difficult to claim that this is an established means for deterioration learning. While it is difficult to accurately detect separation using the image quality from visual inspections, a deterioration learning model can potentially be used for rebar exposure using the image quality from visual inspections. This paper proposes a practical method applies semantic segmentation (segmentation) of concrete damage using images of damage from close eye-base inspections. Results are shown from actually applying this method on sparse images of damage, focusing on images of rebar exposure among images of damage to bridges. Finally, references will be made to issues of damage detection modeling as well.

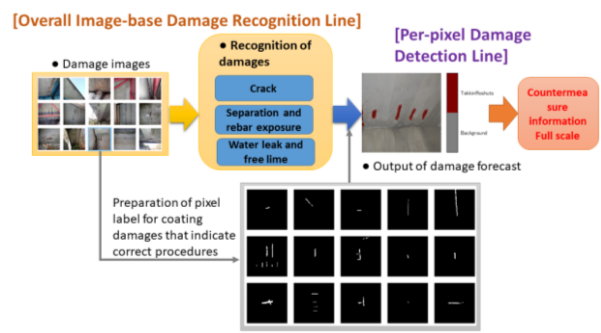

Figure 1: Deterioration learning workflow line from eye-inspected damaged images to per-pixel region prediction

## 2. Related Studies

### 2.1 Damage detection studies for civil infrastructures

Since 2002, there has been an accumulation of studies (Wu, 2002) (Chun, 2015) on resolving damage detection using neural networks (ANN) for the purpose of continuous surveillance of bridges. Many instances of damage detection modeling for machine learning have been conducted over the past 15 years, including the ANN, as well as the PCA, SVM, GA and other such solution methods (Gordan, 2017). Since the potential of convolutional neural networks (CNN) to exhibit high degrees of accuracy in classifying one million images into 1,000 classes was reported in 2012 (AlexNet, 2012), there has been active reporting of studies on solution methods of the CNN, which provides solutions with greater accuracy than conventional methods for label categorization of overall images, object detection and semantic segmentation at the pixel level. There have been a

Yasuno Takato, address: 5-20-8, Asakusabashi, Taito-ku, Tokyo, 111-8648, e-mail: tk-yasuno@yachiyo-eng.co.jp

number of studies conducted on damage classification of at the whole-image level for cracks and corrosion of road pavement, structures and bridges, for detection of damage to civil engineering structures (Gopalakrishnan, 2018) (Ricard, 2018), as well as damage segmentation at the pixel level (Hoskere, 2017). A report was made on a study that applied deep CNN to conduct four classes of damage segmentation, namely no damage, only separation, exposure of rebar (with and without rust), using 734 images of damage (Guillamon, 2018). The breakdown of the damage classes, however, indicated a distribution biased to the third class, for which there were 510 images, and as such, distortion in the training images cannot be denied. Dimensions of the images of damage were widely varied, being 640 x 480, 800 x 600, 1,024 x 768, 1,280 x 960 and 1,600 x 1,200. The potential for learning with the index that represents the degree of matching between prediction and reality, mIoU (class mean Intersection of Union) to the level of 0.6 to 0.8 was indicated by using some types of CNN models for fully convolutional networks (FCN) in entering images of such diverse dimensions. The use of the damage detection modeling that utilizes solution method of CNN, however, has just been started and as such, it would be difficult to claim that this is an established general-purpose method for detection of damage in management of bridges. A practical method for damage segmentation with considerations for characteristics of images of damage from close eye-base inspections of bridges is proposed by this paper as follows.

Table 2: Damaged region of interest (ROI) to background ratio based on pixel counts per bridge inspection images, the ROI is very sparse.

| Example consisting of 208 damage photographs that reveal rebar exposure | Total number of pixels per damage image | Average number of pixels per image | Percentage per image |
|---|---|---|---|
| Background | 63,683,619 | 306,171 | 98.9% |
| **Damage to region of interest (ROI)** | 725,251 | 3,487 | **1.1%** |
| Total per image | 64,408,870 | 309,658 | 100.0% |

## 2.2 Characteristics of eye-inspection images

This paper provides a practical observation on characteristics of images of damage, using 208 images of damage in which exposure of rebar has been captured through close eye-base inspection of bridges, which are intended targets. While generality cannot be guaranteed with these characteristics, they are considered to lead the way for utilizing images of damage. Characteristics of general conditions and damage for separations and rebar exposure are such that the condition in which the surface of the concrete member has separated is referred to as "separation", while the where rebar is exposed in such separated member is referred to as "rebar exposure". The five-level evaluation categories consist of a and b for no damage, c for separation only, d for rebar exposure that is slight in degree and e for rebar exposure with significant corrosion or fracture of rebar (Ministry of Land, Infrastructure, Transport and Tourism, 2014). The summary value for the damage area (region of interest: ROI) subject to detection, as well as other regions in the background, counted at pixel level, is shown in Table 2. No advance manipulation was conducted on images to unify photographing distance and picture quality. The average number of pixels per image was 309 thousand pixels. The proportion of these that include targeted damage was merely 1%. The first characteristic of damage image is the sparsity of the area comprised of ROI.

## 3. Learning Deep Neural Network via Semantic Segmentation toward Damage Images

The FCN-Alex and FCN-VGG16 (Long, 2015), as well as the SegNet-VGG16 (Badrinarayanan, 2016) are compared where appropriate, as a method for learning transfers of semantic segmentation. The solution method used in this paper by itself does not present any innovation but the extremely sparse proportion of detection target ROI on any given image is a characteristic and the intention was to derive a practical method that can be applied to images of damage with sparse pixel labels. The FCN-Alex is a transfer learning of AlexNet and the CNN is implemented to the deepest layer, making it a deep neural net (DNN) of 23 layers in depth. Learning is possible with relatively short calculation time and prediction output for exhaustive detection of targeted damage can be achieved. Next, FCN-VGG16 (16s) is derived by transfer learning of VGG16 and while this method requires a long time for calculation, the DNN has a compact network structure with depth of 47 layers. SegNet-VGG16 is a method of transfer learning used to identify objects for automatic driving and a DNN with depth of 91 layers. Furthermore, we tried to create a DNN using the U-Net (Ronneberger 2015) from a scratch where the depth of encoder-decoder layers is five so the number of layers is 70.

This paper applies the four deep neural networks described above to images of damage to compare calculation execution time, accuracy and prediction output image. There is a problem of no improvements being evident with loss functions when the SGDM is used in the optimization method for hyper-parameters, as gradients of the detection target are eliminated due to the sparse characteristic of the damage image. In order to overcome this issue, the gradient of the detection target is captured with good sensitivity and the previously updated quantities are deleted where appropriate, and the RMSProp, which has a characteristic formula for error function that eliminates the amount of change in gradients of detection targets by taking square root of the amount of change in gradient, is adopted (Hinton, 2012) (Mukkamala, 2017). The weighting factor for the updating amount was set to 0.99. The learning coefficient for the overall model was set to 1E-5 and the minibatch was set to 16.

## 4. Applied Results

### 4.1 Deep learning results

The usage rate of the training and test data for 208 images of damage of rebar exposure from close eye-base inspections of 106 bridges was set to Train: Test = 95:5. The transition of loss function in the learning process applied to the rebar exposure segmentation is shown in Figure 2. The calculation conditions are 832 cycles per epoch for a total of 8,320 repeated calculations in 10 epochs. The loss function of FCN-VGG16 is shown in significant depth, transitioning at a minimum level. The loss value of the FCN-AlexNet is transitioning at a higher level than FCN-VGG16. These two FCN models, however, have large dispersion of loss values and their disadvantage is that they make for unstable learning processes. The loss function of the SegNet-VGG16 does not offer minimum values, but up and down fluctuations remain small early on, which can be interpreted to offer superior stability for the learning process.

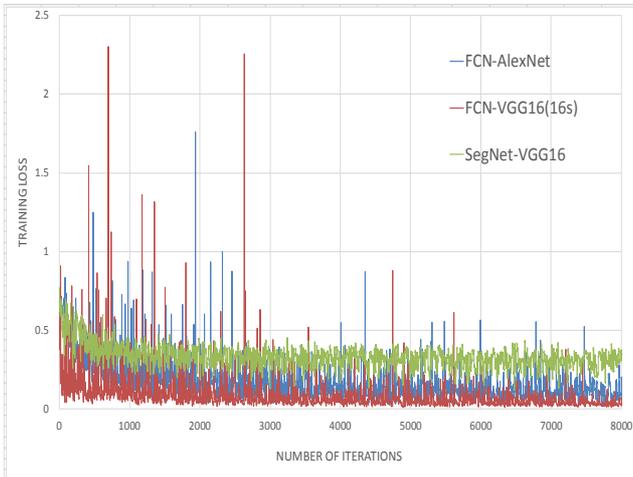

Figure 2: Training process of rebar exposure segmentation loss function

Table 3: Comparison of indices for rebar exposure segmentation models

| DNN model | Time calculation | Average mIoU | Weighted wIoU |
|---|---|---|---|
| FCN-AlexNet | 129 min. | 0.5291 | 0.9639 |
| FCN-VGG16 (16s) | 460 min. | 0.6662 | 0.9775 |
| **SegNet-VGG16** | **230 min.** | **0.7757** | **0.9897** |
| U-Net (depth=5) | 757min. | 0.5099 | 0.9532 |

Calculation time, accuracy, average and weighted IoU index of respective model are shown in Table 3. The FCN-Alex offers a relatively short calculation time of just 129 minutes for learning, but the IoU index is not all that high. The FCN-VGG16 requires the longest calculation time of 7 hours and 40 minutes, but offers a superior IoU index that matches reality with prediction of damage to pixel level. Furthermore, in the predicted output of the U-Net trial with the encoder-depth five, some linear part which is not a reinforcing rebar appears. Since the U-Net is a creation of a new CNN from the scratch, learning iterations has been carried out up to 20 epochs. The U-Net achieved the index such as average mIoU = 0.5099, and weighted wIoU = 0.9532. Thus, the SegNet-VGG16 offers average mIoU of 0.7757 and weighted wIoU of 0.9897, to achieve the highest accuracy as far as we compute these bridge eye-inspection images and annotated labels.

### 4.2 Prediction results

Output of predictions for test images, using the model that involves learning of rebar exposure segmentations as described above are introduced below. Output of segmentation predictions for respective models for test images that depict rebar exposure is shown in Figure 3. Output of predictions for the FCN-AlexNet, FCN-VGG16 (16s), SegNet-VGG16, and U-Net, top to bottom in four levels, are shown. The left side shows the pixel level prediction of the background colored gray, based on the original image, while the pixel level image of rebar exposure prediction is colored in brown. The right side shows pixels of the prediction and reality that match, in white. Pixels that were falsely detected and where prediction does not match reality, are shown in green. Pixels that were not predicted against reality and were therefore omitted, are shown in magenta.

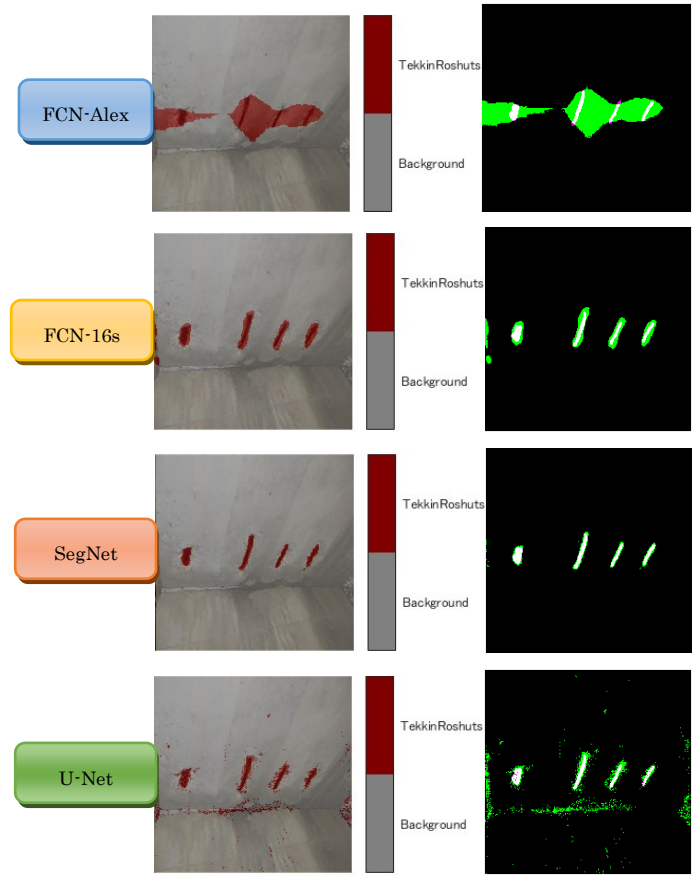

Figure 3: Test image prediction of rebar exposure segmentation by our trained networks (left: prediction (red pixels) over eye-inspection image, right: predicted mask overlaid with the ground truth. Note: white pixels are true damaged, green are over precision, and magenta are less recall.)

The prediction output for the FCN-AlexNet captured the rebar exposure in reality (white) without omission and with high reproducibility. Predictions were also made on the surroundings (green), which increased false detections and thereby lowered accuracy rate. The prediction for the FCN-VGG16 (16s) on the second level likewise captured reality without any omission and improved on the false detection for the surroundings. The prediction output for the SegNet-VGG16 on the third level captured rebar exposure in reality without any omission and offered extremely low false detections for the surroundings to score highest level among the four models for both reproducibility and accuracy rate. Prediction outputs for rebar exposure were verified for ten test images, which constitute 5% of 208 images and same levels of accuracy as described above were obtained. Other prediction outputs will be introduced on the day of presentation, due to the constraints of space on the paper.

## 5. Conclusion

### 5.1 Concluding remarks

This paper proposed a method for detecting rebar exposure by segmentation, using sparse images of damage obtained from close eye-base inspections. Specifically, learning was attempted on four rebar exposure segmentations. This method was actually applied to images of damage with relatively low image quality and size heterogeneity. Annotation images for rebar exposure were prepared and pre-process that involved random generation of patches was implemented to increase the number and variation of images of damage. Learning of high accuracy, based on transfer learning is now possible, even to images that are entered with low image quality and size heterogeneity. FCN-AlexNet and FCN-VGG16 exhibited high reproducibility (recall) by detecting damage without any omissions, but false detections occurred for the surroundings, which then deteriorated the accuracy rate (precision), which remains as an issue. The SegNet-VGG16 exhibited the best accuracy and achieved class average index of 77.57% and weighted index wIoU of 98.97%. The study made it evident that damage segmentation can be incorporated with damage detection modeling, by utilizing transfer learning of images from conventional close human eye-inspections and even without homogenous image capturing conditions with 4k quality.

### 5.2 Future works

Issues for future shall be mentioned. The scope of this paper was the detection of rebar exposure, using images of damage from close eye-base inspection of bridges. The standard for inspection of bridges prescribes 26 items (Ministry of Land, Infrastructure, Transport and Tourism, 2014). Creation of dataset for learning and learning of damage detection models for other types of damage, which are indicative of rebar exposure, such as "separation", "water leak and free lime" and "crack" is the issue. These will involve more details than rebar exposure and are covered by concrete in the background to offer little characteristics. The conventional close eye-base inspection shall therefore be considered the primary screening, while conditional sorting and learning trials for recording more detailed images of damage captured with high quality 4k resolution as localized detailed inspection only on damage that require attention would be an issue. Infrastructure administrators manage many aging structures other than bridges as well. Learning of damage detection models using a diverse range of images of damage for a wide variety of other structures will be the issue for the future. Creation of detection intelligence created from scratch, by accumulating images is also a challenging issue. Per-pixel images are restricted to represent the depth feature of craterous concrete surface. So we will tackle 3D volume segmentation such as point cloud data mining and damage volume prediction.

[Acknowledgments] We wish to thank Shinichi Kuramoto and Takuji Fukumoto to provide us practical MATLAB information.

(e-print submitted April 22, 2020)